\pdfoutput=1

\documentclass[11pt]{article}

\usepackage[table]{xcolor}
\usepackage{colortbl}
\usepackage{geometry}
\usepackage{array}
\usepackage{booktabs}
\usepackage{caption}
\usepackage{multirow}
\usepackage{pifont} %
\usepackage{hyperref}
\usepackage{svg}
\usepackage{colortbl}
\usepackage{amsmath}
\usepackage{tcolorbox}
\usepackage{amssymb}
\usepackage{pifont}
\usepackage{mdframed}
\usepackage{balance}
\newcommand{\xmark}{\ding{55}}
\newcommand{\negshot}{PAP}
\newcommand{\negnum}{\emph{Neg-Num}}
\newcommand{\approximation}{\emph{Approx}}
\newcommand{\numeration}{\emph{Num}}
\newcommand{\random}{\emph{Rand-Repl}}
\newcommand{\mask}{\emph{Mask}}
\newcommand{\range}{\emph{Range}}
\newcommand{\gemini}{Gemini 2.5F}
\newcommand{\geminiT}{Gemini 2.5F$^T$}
\newcommand{\deepseek}{DeepSeek-R1}
\newcommand{\deepseekT}{DeepSeek-R1$^T$}
\newcommand{\qwen}{Qwen3-32B}
\newcommand{\qwenT}{Qwen3-32B$^T$}
\newcommand{\gptoThree}{GPT-o3$^T$}
\newcommand{\gptFiveT}{GPT-5$^T$}
\newcommand{\llamaThree}{Llama 3.3-70B}
\newcommand{\gptFourO}{GPT-4o}
\newcommand{\gptFourOMini}{GPT-4o-Mini}

\definecolor{lightred}{RGB}{255,230,230}
\definecolor{lightyellow}{RGB}{255,255,230}
\definecolor{lightgreen}{RGB}{230,255,230}
\definecolor{lightgrey}{RGB}{235,235,235}
\definecolor{darkgreen}{RGB}{0,100,0}

\usepackage[final]{acl}

\usepackage{times}
\usepackage{latexsym}

\usepackage[T1]{fontenc}

\usepackage[utf8]{inputenc}

\usepackage{microtype}

\usepackage{inconsolata}

\usepackage{graphicx}
\usepackage{enumitem}
\usepackage{subcaption}
\usepackage{todonotes}

\title{NumPert: Numerical Perturbations to Probe Language Models for Veracity Prediction}

\author{
  \textbf{Peter Røysland Aarnes} \quad \textbf{Vinay Setty} \\
  University of Stavanger \\
  \texttt{peter.r.aarnes@uis.no, vsetty@acm.org}
}

\begin{document}
\maketitle

\begin{abstract}
Large language models show strong performance on knowledge intensive tasks such as fact-checking and question answering, yet they often struggle with numerical reasoning. We present a systematic evaluation of state-of-the-art models for veracity prediction on numerical claims and evidence pairs using controlled perturbations, including label-flipping probes, to test robustness. Our results indicate that even leading proprietary systems experience accuracy drops of up to 62\% under certain perturbations. No model proves to be robust across all conditions. We further find that increasing context length generally reduces accuracy, but when extended context is enriched with perturbed demonstrations, most models substantially recover. These findings highlight critical limitations in numerical fact-checking and suggest that robustness remains an open challenge for current language models.
\end{abstract}

\section{Introduction}

\begin{figure}[t!]

\centering
\begin{tcolorbox}[title= Numerical Perturbation Example,
left=1pt, colback=white]

\footnotesize
\begin{tabular}{p{\columnwidth}}

\textbf{Original Claim:} \emph{``In 2020, the company's revenue was \textcolor{darkgreen}{5,000,000} dollars, making a significant growth from the previous year''.} \\

[\textbf{Label:} \texttt{TRUE}, \textbf{Model Prediction: \textcolor{darkgreen}{\texttt{TRUE}}}~$\checkmark$] \\[0.5em]

\textbf{Perturbed Claim:} \emph{``In 2020, the company's revenue was \textcolor{red}{fifty million} dollars
making a significant growth from the previous year.''} \\[0.3em]

[\textbf{Label}: \texttt{FALSE},  \textbf{Model Prediction: \textcolor{red}{\texttt{TRUE}}}~\xmark] \\[0.5em]

\textbf{Evidence:} \emph{``A market analysis by MNO Research Group, published in 2021,
states: 'PQR Innovations experienced significant growth (...). The revenue for the year 2020 reached \textcolor{darkgreen}{5,000,000} dollars.''}[1em]
\vspace{-5pt}
\end{tabular}
\end{tcolorbox}
\vspace{-10pt}
\caption{\small Example illustrating how the original `TRUE' claim is perturbed into a `FALSE' claim, yet the model predicts `TRUE'.}
\vspace{-20pt}
\label{fig:numeration_example}
\end{figure}

Verifying claims in social media, political debates, and press releases has become essential. 
While platforms such as Politifact, Snopes, and FullFact support manual fact-checking, their scalability is limited. 
Numerical claims, in particular, are tedious and error prone for human annotators~\cite{Aly:2021:NURIPS}. 
Neural language models provide a promising alternative for evidence retrieval and preliminary veracity assessment~\cite{Guo:2022:TACL,Dmonte:2024:arXiv,Setty:2024:SIGIR}. 
Yet, recent studies show that both transformer models fine-tuned for numerical claim verification and general purpose large language models struggle with numerical reasoning~\cite{Wallat:2024:WSDM,V:2024:Sigir,Akhtar:2023:ACL}, and the reasons remain unclear.

Although prior work has studied LLM fragility in numerical reasoning for QA~\cite{Xu:2022:EMNLP} and tabular NLI~\cite{Akhtar:2023:ACL}, no systematic analysis exists for veracity prediction in long-context fact-checking. Our results indicate that models are prone to errors with longer context and reasoning chains. To address this gap, we evaluate state-of-the-art models of different sizes and architectures under varied prompting settings with systematically perturbed numerical claims and evidence.

Manipulating numerical values in unstructured text requires care to ensure that perturbations remain meaningful. We define six probe types: Numeration (\numeration), Approximation (\approximation), \range, Masking (\mask), Random Replacement (\random), and Negative Number (\negnum) (see Table~\ref{tab:transformation_matrix}) to systematically modify numbers while preserving claim intent. In some cases, these perturbations also flip the factual label (e.g., changing \$5{,}000{,}000 to fifty million; see Figure~\ref{fig:numeration_example}). All perturbations are manually verified to ensure correctness and relevance. This study addresses three research questions: 
\begin{description}[leftmargin=0pt,labelindent=0pt]
    \item[\textbf{RQ1}]: \textit{Which models in our selection of diverse sizes are most and least robust?} 
    \item[\textbf{RQ2}]: \textit{Which numerical perturbations most affect performance?} 
    \item[\textbf{RQ3}]: \textit{How do context length and reasoning chains influence robustness?}
\end{description}
To answer this, we test models on claim–evidence pairs, comparing baseline predictions with those on numerically perturbed claims. Larger gaps reflect weaker robustness. We use truthful probes that keep the original label and label-flipping probes that contradict the evidence, under zero-shot, two-shot, and perturbation aware prompts (\negshot).

Our results show that all state-of-the-art models are highly vulnerable to numerical perturbations, particularly under \mask~and \negnum. We also notice that zero-shot settings outperform two-shot, while providing a few perturbed examples (\negshot~prompt) helps models recover in most cases. These findings reveal weaknesses in LLM veracity prediction.

\section{Related work}

The interpretability of LLMs is critical for knowledge-intensive tasks like question answering and fact-checking. Probing studies have revealed their opaque decision processes \cite{Belinkov:2022:CL}. For instance, \citet{Yang:2024:arXiv,Lu:2023:ACL,Frieder:2024:NeurIPs} showed that while LLMs can perform complex reasoning, they often struggle with basic numeracy.

Several works have examined numerical reasoning in LLMs. \citet{Wallace:2019:EMNLP} probed embeddings from BERT and GloVe, finding inherent but inconsistent numeracy. \citet{Akhtar:2023:ACL} evaluated models on tabular data with a hierarchical taxonomy, showing no model excels across all tasks. \citet{Xu:2022:EMNLP,Zhou:2024:ACL} demonstrated that numerical perturbations in QA often mislead LLMs, while \citet{Paruchuri:2024:EMNLP,Chen:2024:arXiv}  highlighted weaknesses in numerical reasoning. Several studies also reveal that LLMs for fact-checking are brittle to textual perturbations, and adversarial edits~\cite{Mamta:2025:NAACL,Przybyla:2024:NLPJ,Liu:2025:Survey}.

Despite prior advances, key gaps remain. Most work does not examine numerical reasoning in \emph{open-domain fact-checking} with real-world, long-context data, and reproducibility is often limited. For instance, \citet{Akhtar:2023:ACL} rely on synthetic tabular inputs with short context, provide incomplete perturbation details, and lack an accessible repository. In contrast, we evaluate numerical reasoning in realistic, unstructured settings, introduce perturbations that preserve semantic validity, and release full code and data to ensure reproducibility.

\section{Methodology}
Our study examines veracity prediction models by systematically perturbing numerical values in claims to assess their impact on label prediction. The methodology involves (1) curating a dataset with diverse numerical expressions (e.g., statistics), (2) applying controlled perturbations (e.g., scaling, replacements, masking). (3) Extensive error analysis leveraging the reasoning tokens.
\newcommand{\cmark}{\ding{51}} %

\begin{table}[ht!!]
\small
\centering
\caption{\small The number of  claims per perturbation type  that remain `True' (T$\rightarrow$T), remain `False' (F$\rightarrow$F), or switch from `True' to `False' (T$\rightarrow$F). Unperturbed baseline has 260 \texttt{True} claims and 604 \texttt{False} claims.}
\label{tab:transformation_matrix}
\begin{tabular}{lcccc}

\toprule
\textbf{Category} & \textbf{T $\rightarrow$ T} & \textbf{F $\rightarrow$ F} & \textbf{T $\rightarrow$ F}  \\
\midrule
\textbf{\numeration} & 213   &490 & 213 \\
\textbf{\approximation} & 170 & 404 & 170 \\
\textbf{\range} & 188 & 411 & 188 \\
\textbf{\mask} & \xmark & 490 & 213 \\
\textbf{\random} & \xmark & 490 & 213 \\
\textbf{\negnum} & \xmark & 89 & 51 \\
\bottomrule
\end{tabular}
\end{table}

\subsection{Dataset and Preprocessing}
We use the \textit{QuanTemp}  dataset~\cite{V:2024:Sigir}, which contains real world claim-evidence pairs with numerical focus from reputable fact checking sources. Each pair is labeled as \texttt{True}, \texttt{False}, or \texttt{Conflicting}. For our evaluation, we exclude the \textit{Conflicting} class due to its inherent ambiguity.  To prevent shortcut learning, we remove summaries from all pairs, requiring models to assess veracity solely from evidence.  

Each claim is processed with the spaCy NER tagger (covering  \textit{Cardinal}, \textit{Money}, \textit{Percent}, \textit{Time}, \textit{Date}, and \textit{Ordinal}), and numerical values are normalized to digits using the \textit{Word2Number} library (similar to \cite{Akhtar:2023:ACL,Wallace:2019:EMNLP,Xu:2022:EMNLP}). Perturbed claims are \textit{manually verified} for validity, and invalid cases are removed.

\subsection{Perturbation Techniques}

We adopt the numerical reasoning taxonomy of \citet{Akhtar:2023:ACL} (see Table~\ref{tab:transformation_matrix}). The \numeration, \approximation, and \range~settings perturb numbers while remaining consistent with the evidence, so \texttt{True} claims stay \texttt{True}. Conversely, \mask, \random, and \negnum~modify values such that \texttt{True} claims flip to \texttt{False}, while \texttt{False} claims remain unchanged. We do not perturb \texttt{False} to \texttt{True}, since falsity can stem from multiple factors and counterfactual claims are often infeasible. Exploring this direction is left for future work. Now we explain the different perturbation techniques:

\noindent \textbf{\numeration:} Tests whether models recognize equivalence between digits and words (e.g., ``12'' vs.\ ``twelve''), preserving the original label for non-flipping probes. Perturbation applies to \texttt{Cardinal}, \texttt{Percent}, and \texttt{Money}, but not to \texttt{Ordinal}, \texttt{Time}, or \texttt{Date}, except for cardinal numbers within \texttt{Time} (e.g., ``24 hours'' to ``twenty four hours'').  For the label-flipping probes, the original number is modified (e.g., ``12'' could be perturbed to ``fifteen'').

\noindent \textbf{\approximation:} Non-flipping probes reduces precision by rounding and adding \emph{about} (e.g., ``1,025 dollars'' to ``about 1000 dollars''), retaining truth when close to the evidence.  For the label-flipping probes, the original value is altered so that it is no longer reflective of the true amount (e.g., original``1,025 dollars'' to ``about 1200 dollars'').

\noindent \textbf{\range:} Non-flipping probes replaces exact values with spans (e.g., ``25 percent'' to ``between 20 and 30 percent''), testing reasoning over intervals. The label-flipping probes modifies the span such that the original number is not within it (e.g., the original ``25 percent'' is perturbed to ``between 30 and 40 percent'').

\noindent \textbf{\random:} Replaces numbers with random values of equal digit length (e.g., ``100,000'' to ``423,823''), mismatching the evidence.  

\noindent \textbf{\mask:} Hides numbers with ``\#'' tokens according to digit length, including delimiters (e.g., ``100,000'' to ``\#\#\#\#\#\#\#''), requiring inference from evidence.  

\noindent \textbf{\negnum:} Converts values to negatives (e.g., ``4\%'' to ``-4\%''), applied only to percentages since other entities (money, time, dates) typically use linguistic cues like ``decrease.''  

\subsection{Prompting Strategy}
\label{subsec:prompt}
All models use identical instructions under three prompting strategies: (1) \textbf{Zero-shot} with only instructions and no demonstrations (see Appendix \ref{app:prompt}), (2) \textbf{Two-shot} prompt that extends the zero-shot prompt with one \texttt{True} and one \texttt{False} demonstration from  training data with evidence and rationale~\cite{Brown:2020:NeurIPS}. (3) We also test models with a perturbation aware prompt \textbf{(\negshot)}, which pairs a perturbed claim with one sentence evidence for each perturbation type and flipped label. A similar approach is used by \cite{Hu:2024:SIGKDD} in a RAG setting. Full prompts are provided in Appendix~\ref{app:prompt}.

\section{Experimental Setup}
This section describes our experimental framework, including the language models used, and evaluation methods.

\subsection{Model Selection}

\begin{description}[leftmargin=0.2cm, labelindent=0cm, nosep]

 \item[\emph{Open‑weight LLMs:}] \emph{DeepSeek‑R1‑32B}, \emph{Qwen3‑32B}, \emph{Llama3.3‑70B}, \emph{Llama 3.2‑1B}, and \emph{Mistral‑7B} (All models are from Ollama framework\footnote{\url{https://ollama.com/search}} with Q4\_K\_M quantization).
 
 \item[\emph{Proprietary LLMs:}] \emph{GPT‑4o~(v2024-08-06)}, \emph{GPT‑4o‑mini (v2024-07-18)}, \emph{GPT‑5 (v2025-08-07)}, \emph{GPT‑o3 (v2025-04-16)}, and \emph{Gemini 2.5 Flash (v2025-06)} (All models are accessed via their respective official APIs)
\end{description}

Models with thinking are marked with superscript $T$. All models ran with temperature 0 and JSON output; open-weight and OpenAI models used default (medium) reasoning effort. For Gemini 2.5 Flash$^T$, we fixed the thinking budget to 8192 (vs. the default 1) for cost efficiency. Other settings followed defaults. We exclude \emph{Llama 3.2-1B} and \emph{Mistral-7B} from the main results due to limited robustness; details are in Appendix~\ref{app:f2f}. Invalid predictions are rare, except for \deepseekT, which yields 6.8\% invalid outputs under zero-shot. Thinking variants generally produce more invalid outputs than their non-thinking counterparts (see Appendix~\ref{sec:invalid_analysis}). Code and data can be accessed though our GitHub repository\footnote{\url{https://github.com/iai-group/adversarial_attack_numerical_claims/}}.

\subsection{Evaluation}
Robustness is assessed by comparing baseline performance on non-perturbed claims with performance on perturbed ones. We use per-class accuracy metric. We use accuracy as the primary metric for $T\rightarrow F$ evaluations. To gain greater insight into model errors, we manually analyze reasoning tokens of zero-shot vs. \negshot~for \texttt{T} $\rightarrow$ \texttt{F} claims to look for common patterns that models fall into while evaluating a claim.

\section{Results}
\label{results_section}
We report results across models and perturbation settings. We first describe performance on unperturbed claims, then analyze changes under non-flipped and flipped label conditions. Results for \texttt{False} $\rightarrow$ \texttt{False} cases are omitted here for brevity (see Appendix~\ref{app:f2f}). Models are evaluated under three prompting regimes defined in Section \ref{subsec:prompt} (see Appendix~\ref{app:prompt} for full prompts).

\begin{table*}[h!]
\centering
\small
\caption{Accuracy (reported in \%) for `True' dataset split for label flips perturbations (True $\rightarrow$ False), and comparing accuracy variance between the flipped probes to model performance on unaltered \textit{original} claims accuracy (\textcolor{red}{-x} indicates a drop; \textcolor{darkgreen}{+x} indicates an increase). Values in bold denote the highest accuracy within each perturbation setting, separated by open-weight and proprietary models.} 
\label{tab:t_to_f_results}
\begin{tabular}{lccccccc}

\hline
\textbf{Model} & \textbf{Original} & \textbf{Approx} & \textbf{Neg-num} & \textbf{Num} & \textbf{Rand-repl} & \textbf{Range} & \textbf{Mask} \\
\midrule
\multicolumn{8}{c}{\textbf{Zero-shot}}  \\
\midrule
Llama3.3-70B & 87.32 & 87.65\textsuperscript{\textcolor{darkgreen}{+0.32}} & \textbf{62.75}\textsuperscript{\textcolor{red}{-24.58}} & 68.54\textsuperscript{\textcolor{red}{-18.78}} & \textbf{91.08}\textsuperscript{\textcolor{darkgreen}{+3.76}} & 82.45\textsuperscript{\textcolor{red}{-4.88}} & 10.80\textsuperscript{\textcolor{red}{-76.53}} \\
DeepSeek-R1-32B & 81.69 & \textbf{89.41}\textsuperscript{\textcolor{darkgreen}{+7.72}} & 39.22\textsuperscript{\textcolor{red}{-42.47}} & 56.34\textsuperscript{\textcolor{red}{-25.35}} & 88.73\textsuperscript{\textcolor{darkgreen}{+7.04}} & 81.91\textsuperscript{\textcolor{darkgreen}{+0.22}} & \textbf{23.47}\textsuperscript{\textcolor{red}{-58.22}} \\
DeepSeek-R1-32B$^T$ & \textbf{87.44} & 85.06\textsuperscript{\textcolor{red}{-2.37}} & 31.91\textsuperscript{\textcolor{red}{-55.52}} & 69.43\textsuperscript{\textcolor{red}{-18.01}} & 84.73\textsuperscript{\textcolor{red}{-2.71}} & 86.98\textsuperscript{\textcolor{red}{-0.46}} & 10.63\textsuperscript{\textcolor{red}{-76.81}} \\
Qwen3-32B & 84.35 & 78.24\textsuperscript{\textcolor{red}{-6.12}} & 43.14\textsuperscript{\textcolor{red}{-41.21}} & 58.78\textsuperscript{\textcolor{red}{-25.57}} & 84.51\textsuperscript{\textcolor{darkgreen}{+0.16}} & 80.32\textsuperscript{\textcolor{red}{-4.03}} & 16.43\textsuperscript{\textcolor{red}{-67.92}} \\
Qwen3-32B$^T$ & 85.99 & 89.38\textsuperscript{\textcolor{darkgreen}{+3.38}} & 34.04\textsuperscript{\textcolor{red}{-51.95}} & \textbf{78.24}\textsuperscript{\textcolor{red}{-7.75}} & 87.88\textsuperscript{\textcolor{darkgreen}{+1.89}} & \textbf{87.64}\textsuperscript{\textcolor{darkgreen}{+1.65}} & 12.38\textsuperscript{\textcolor{red}{-73.61}} \\
\hline
GPT-4o & 80.00 & 88.82\textsuperscript{\textcolor{darkgreen}{+8.82}} & 47.06\textsuperscript{\textcolor{red}{-32.94}} & 73.24\textsuperscript{\textcolor{red}{-6.76}} & 90.61\textsuperscript{\textcolor{darkgreen}{+10.61}} & 91.49\textsuperscript{\textcolor{darkgreen}{+11.49}} & 19.25\textsuperscript{\textcolor{red}{-60.75}} \\
GPT-4o-Mini & \textbf{85.38} & 68.24\textsuperscript{\textcolor{red}{-17.15}} & 25.49\textsuperscript{\textcolor{red}{-59.89}} & 56.81\textsuperscript{\textcolor{red}{-28.58}} & 78.87\textsuperscript{\textcolor{red}{-6.51}} & 75.00\textsuperscript{\textcolor{red}{-10.38}} & 11.27\textsuperscript{\textcolor{red}{-74.12}} \\
GPT-5$^T$ & 76.15 & 93.53\textsuperscript{\textcolor{darkgreen}{+17.38}} & 33.33\textsuperscript{\textcolor{red}{-42.82}} & \textbf{86.38}\textsuperscript{\textcolor{darkgreen}{+10.23}} & 89.20\textsuperscript{\textcolor{darkgreen}{+13.05}} & 92.02\textsuperscript{\textcolor{darkgreen}{+15.87}} & 19.72\textsuperscript{\textcolor{red}{-56.44}} \\
GPT-o3$^T$ & 75.77 & 89.41\textsuperscript{\textcolor{darkgreen}{+13.64}} & 25.49\textsuperscript{\textcolor{red}{-50.28}} & 84.98\textsuperscript{\textcolor{darkgreen}{+9.21}} & 88.73\textsuperscript{\textcolor{darkgreen}{+12.96}} & 90.96\textsuperscript{\textcolor{darkgreen}{+15.19}} & 21.60\textsuperscript{\textcolor{red}{-54.17}} \\
Gemini 2.5F & 82.69 & \textbf{95.29}-\textsuperscript{\textcolor{darkgreen}{+12.60}} & 54.90\textsuperscript{\textcolor{red}{-27.79}} & 83.57\textsuperscript{\textcolor{darkgreen}{+0.88}} & \textbf{96.71}\textsuperscript{\textcolor{darkgreen}{+14.02}} & \textbf{93.09}\textsuperscript{\textcolor{darkgreen}{+10.39}} & \textbf{25.82}\textsuperscript{\textcolor{red}{-56.87}} \\
Gemini 2.5F$^T$ & 71.54 & 88.82\textsuperscript{\textcolor{darkgreen}{+17.29}} & \textbf{58.82}\textsuperscript{\textcolor{red}{-12.71}} & 82.63\textsuperscript{\textcolor{darkgreen}{+11.09}} & 89.67\textsuperscript{\textcolor{darkgreen}{+18.13}} & 90.43\textsuperscript{\textcolor{darkgreen}{+18.89}} & 16.90\textsuperscript{\textcolor{red}{-54.64}} \\
\midrule
\multicolumn{8}{c}{\textbf{Two-shot}}  \\
\midrule
Llama3.3-70B & \textbf{91.55} & 72.35\textsuperscript{\textcolor{red}{-19.20}} & \textbf{33.33}\textsuperscript{\textcolor{red}{-58.22}} & 46.48\textsuperscript{\textcolor{red}{-45.07}} & 78.26\textsuperscript{\textcolor{red}{-13.29}} & 57.98\textsuperscript{\textcolor{red}{-33.57}} & 8.92\textsuperscript{\textcolor{red}{-82.63}} \\
DeepSeek-R1-32B & 89.67 & 65.29\textsuperscript{\textcolor{red}{-24.38}} & 21.57\textsuperscript{\textcolor{red}{-68.10}} & 37.09\textsuperscript{\textcolor{red}{-52.58}} & 74.70\textsuperscript{\textcolor{red}{-14.97}} & 58.51\textsuperscript{\textcolor{red}{-31.16}} & 12.21\textsuperscript{\textcolor{red}{-77.46}} \\
DeepSeek-R1-32B$^T$ & 86.32 & 88.55\textsuperscript{\textcolor{darkgreen}{+2.23}} & 22.00\textsuperscript{\textcolor{red}{-64.32}} & 71.15\textsuperscript{\textcolor{red}{-15.17}} & 88.49\textsuperscript{\textcolor{darkgreen}{+2.17}} & 87.17\textsuperscript{\textcolor{darkgreen}{+0.85}} & 9.43\textsuperscript{\textcolor{red}{-76.89}} \\
Qwen3-32B & 79.81 & 70.59\textsuperscript{\textcolor{red}{-9.22}} & 37.25\textsuperscript{\textcolor{red}{-42.56}} & 49.77\textsuperscript{\textcolor{red}{-30.05}} & 66.40\textsuperscript{\textcolor{red}{-13.41}} & 72.87\textsuperscript{\textcolor{red}{-6.94}} & \textbf{20.66}\textsuperscript{\textcolor{red}{-59.15}} \\
Qwen3-32B$^T$ & 83.49 & \textbf{86.98}\textsuperscript{\textcolor{darkgreen}{+3.49}} & 27.45\textsuperscript{\textcolor{red}{-56.04}} & \textbf{78.20}\textsuperscript{\textcolor{red}{-5.29}} & \textbf{88.76}\textsuperscript{\textcolor{darkgreen}{+5.26}} & \textbf{87.23}\textsuperscript{\textcolor{darkgreen}{+3.74}} & 12.74\textsuperscript{\textcolor{red}{-70.75}} \\
\hline
GPT-4o & 86.54 & 82.35\textsuperscript{\textcolor{red}{-4.19}} & 33.33\textsuperscript{\textcolor{red}{-53.21}} & 68.54\textsuperscript{\textcolor{red}{-17.99}} & 87.32\textsuperscript{\textcolor{darkgreen}{+0.79}} & 85.64\textsuperscript{\textcolor{red}{-0.90}} & 13.62\textsuperscript{\textcolor{red}{-72.92}} \\
GPT-4o-Mini & \textbf{89.62} & 67.06\textsuperscript{\textcolor{red}{-22.56}} & 27.45\textsuperscript{\textcolor{red}{-62.16}} & 50.70\textsuperscript{\textcolor{red}{-38.91}} & 77.46\textsuperscript{\textcolor{red}{-12.15}} & 73.94\textsuperscript{\textcolor{red}{-15.68}} & 20.19\textsuperscript{\textcolor{red}{-69.43}} \\
GPT-5$^T$ & 77.69 & \textbf{91.18}\textsuperscript{\textcolor{darkgreen}{+13.48}} & 29.41\textsuperscript{\textcolor{red}{-48.28}} & 84.04\textsuperscript{\textcolor{darkgreen}{+6.35}} & 88.26\textsuperscript{\textcolor{darkgreen}{+10.57}} & 88.83\textsuperscript{\textcolor{darkgreen}{+11.14}} & 18.78\textsuperscript{\textcolor{red}{-58.91}} \\
GPT-o3$^T$ & 75.77 & 89.41\textsuperscript{\textcolor{darkgreen}{+13.64}} & 23.53\textsuperscript{\textcolor{red}{-52.24}} & \textbf{85.45}\textsuperscript{\textcolor{darkgreen}{+9.68}} & 89.67\textsuperscript{\textcolor{darkgreen}{+13.90}} & \textbf{90.43}\textsuperscript{\textcolor{darkgreen}{+14.66}} & 22.07\textsuperscript{\textcolor{red}{-53.70}} \\
Gemini 2.5F & 85.00 & 87.06\textsuperscript{\textcolor{darkgreen}{+2.06}} & 35.29\textsuperscript{\textcolor{red}{-49.71}} & 70.89\textsuperscript{\textcolor{red}{-14.11}} & \textbf{94.37}\textsuperscript{\textcolor{darkgreen}{+9.37}} & 85.64\textsuperscript{\textcolor{darkgreen}{+0.64}} & \textbf{22.90}\textsuperscript{\textcolor{red}{-62.10}} \\
Gemini 2.5F$^T$ & 74.23 & 90.00\textsuperscript{\textcolor{darkgreen}{+15.77}} & \textbf{52.94}\textsuperscript{\textcolor{red}{-21.29}} & 82.16\textsuperscript{\textcolor{darkgreen}{+7.93}} & 92.02\textsuperscript{\textcolor{darkgreen}{+17.79}} & 88.83\textsuperscript{\textcolor{darkgreen}{+14.60}} & 15.96\textsuperscript{\textcolor{red}{-58.27}} \\
\midrule
\multicolumn{8}{c}{\textbf{Perturbation Aware Prompt (\negshot)}}  \\
\midrule
Qwen3-32B & 79.34 & 89.41\textsuperscript{\textcolor{darkgreen}{+10.07}} & 76.47\textsuperscript{\textcolor{red}{-2.87}} & 73.71\textsuperscript{\textcolor{red}{-5.63}} & 90.61\textsuperscript{\textcolor{darkgreen}{+11.27}} & 89.36\textsuperscript{\textcolor{darkgreen}{+10.02}} & \textbf{67.61}\textsuperscript{\textcolor{red}{-11.74}} \\
Qwen3-32B$^T$ & 71.23 & \textbf{95.27}\textsuperscript{\textcolor{darkgreen}{+24.04}} & 74.00\textsuperscript{\textcolor{darkgreen}{+2.77}} & \textbf{90.14}\textsuperscript{\textcolor{darkgreen}{+18.91}} & \textbf{94.37}\textsuperscript{\textcolor{darkgreen}{+23.14}} & \textbf{94.62}\textsuperscript{\textcolor{darkgreen}{+23.40}} & 44.85\textsuperscript{\textcolor{red}{-26.38}} \\

\hline
Gemini 2.5F & \textbf{81.92} & \textbf{97.06}\textsuperscript{\textcolor{darkgreen}{+15.14}} & 74.51\textsuperscript{\textcolor{red}{-7.41}} & 84.98\textsuperscript{\textcolor{darkgreen}{+3.05}} & \textbf{97.18}\textsuperscript{\textcolor{darkgreen}{+15.26}} & \textbf{94.68}\textsuperscript{\textcolor{darkgreen}{+12.76}} & 29.11\textsuperscript{\textcolor{red}{-52.82}} \\
Gemini 2.5F$^T$ & 63.08 & 91.76\textsuperscript{\textcolor{darkgreen}{+28.69}} & \textbf{88.24}\textsuperscript{\textcolor{darkgreen}{+25.16}} & \textbf{86.85}\textsuperscript{\textcolor{darkgreen}{+23.78}} & 92.02\textsuperscript{\textcolor{darkgreen}{+28.94}} & 90.96\textsuperscript{\textcolor{darkgreen}{+27.88}} & 26.29\textsuperscript{\textcolor{red}{-36.79}} \\

\hline
\midrule
\end{tabular}
\vspace{-10pt}
\end{table*}

\begin{table}[h!!]
\centering
\small
\caption{Accuracy (reported in \%) on the `True' dataset split under non label-flipping perturbations (True $\rightarrow$ True). The table compares perturbed accuracy to unaltered \textit{original} claim accuracy (\textcolor{red}{-x} indicates a drop; \textcolor{darkgreen}{+x} indicates an increase). Values in bold denote the highest accuracy within each perturbation setting, separated by open-weight and proprietary models.}
\label{tab:t_to_t_results}
\scalebox{0.95}{
\begin{tabular}{lccc}
\hline
\textbf{Model} & \textbf{Approx} & \textbf{Num} & \textbf{Range}  \\
\midrule
\multicolumn{4}{c}{\textbf{Zero-shot}}  \\
\midrule
Llama3.3-70B & 71.76\textsuperscript{\textcolor{red}{-15.56}}  & \textbf{86.38}\textsuperscript{\textcolor{red}{-0.94}}  & 70.21\textsuperscript{\textcolor{red}{-17.11}}  \\
DeepSeek-R1 & 75.29\textsuperscript{\textcolor{red}{-6.40}}  & 82.63\textsuperscript{\textcolor{darkgreen}{+0.94}}  & 67.55\textsuperscript{\textcolor{red}{-14.14}}  \\
DeepSeek-R1$^T$ & \textbf{81.44}\textsuperscript{\textcolor{red}{-6.00}}  & 84.62\textsuperscript{\textcolor{red}{-2.82}}  & \textbf{79.23}\textsuperscript{\textcolor{red}{-8.20}}  \\
Qwen3 & 73.53\textsuperscript{\textcolor{red}{-10.82}}  & 85.88\textsuperscript{\textcolor{darkgreen}{+1.53}}  & 62.23\textsuperscript{\textcolor{red}{-22.12}}  \\
\qwenT & 79.39\textsuperscript{\textcolor{red}{-6.60}}  & 85.02\textsuperscript{\textcolor{red}{-0.97}}  & 78.24\textsuperscript{\textcolor{red}{-7.76}}  \\
\hline
GPT-4o & 68.82\textsuperscript{\textcolor{red}{-11.18}}  & 80.28\textsuperscript{\textcolor{darkgreen}{+0.28}}  & 55.32\textsuperscript{\textcolor{red}{-24.68}}  \\
GPT-4o-Mini & \textbf{81.18}\textsuperscript{\textcolor{red}{-4.21}}  & \textbf{92.96}\textsuperscript{\textcolor{darkgreen}{+7.57}}  & \textbf{79.79}\textsuperscript{\textcolor{red}{-5.60}}  \\
\gptFiveT & 75.29\textsuperscript{\textcolor{red}{-0.86}}  & 77.00\textsuperscript{\textcolor{darkgreen}{+0.84}}  & 73.40\textsuperscript{\textcolor{red}{-2.75}}  \\
\gptoThree & 74.71\textsuperscript{\textcolor{red}{-1.06}}  & 77.46\textsuperscript{\textcolor{darkgreen}{+1.70}}  & 77.66\textsuperscript{\textcolor{darkgreen}{+1.89}}  \\
Gemini 2.5F & 60.69\textsuperscript{\textcolor{red}{-22.00}}  & 79.81\textsuperscript{\textcolor{red}{-2.88}}  & 43.92\textsuperscript{\textcolor{red}{-38.78}}  \\
Gemini 2.5F$^T$ & 68.24\textsuperscript{\textcolor{red}{-3.30}}  & 71.76\textsuperscript{\textcolor{darkgreen}{+0.22}}  & 61.70\textsuperscript{\textcolor{red}{-9.84}}  \\
\midrule
\multicolumn{4}{c}{\textbf{Two-shot}}  \\
\midrule
Llama3.3-70B & 84.71\textsuperscript{\textcolor{red}{-6.84}}  & 90.14\textsuperscript{\textcolor{red}{-1.41}}  & 85.64\textsuperscript{\textcolor{red}{-5.91}}  \\
DeepSeek-R1 & \textbf{88.82}\textsuperscript{\textcolor{red}{-0.85}}  & \textbf{90.61}\textsuperscript{\textcolor{darkgreen}{+0.94}}  & \textbf{86.70}\textsuperscript{\textcolor{red}{-2.97}}  \\
DeepSeek-R1$^T$ & 82.25\textsuperscript{\textcolor{red}{-4.07}}  & 87.50\textsuperscript{\textcolor{darkgreen}{+1.18}}  & 77.13\textsuperscript{\textcolor{red}{-9.19}}  \\
Qwen3-32B & 72.94\textsuperscript{\textcolor{red}{-6.87}}  & 81.69\textsuperscript{\textcolor{darkgreen}{+1.88}}  & 67.02\textsuperscript{\textcolor{red}{-12.79}}  \\
Qwen3-32B$^T$ & 81.66\textsuperscript{\textcolor{red}{-1.83}}  & 84.43\textsuperscript{\textcolor{darkgreen}{+0.94}}  & 77.72\textsuperscript{\textcolor{red}{-5.77}}  \\
\hline
GPT-4o & 77.06\textsuperscript{\textcolor{red}{-9.48}}  & 85.92\textsuperscript{\textcolor{red}{-0.62}}  & 63.83\textsuperscript{\textcolor{red}{-22.71}}  \\
GPT-4o-Mini & \textbf{81.18}\textsuperscript{\textcolor{red}{-8.44}}  & \textbf{89.67}\textsuperscript{\textcolor{darkgreen}{+0.06}}  & \textbf{76.06}\textsuperscript{\textcolor{red}{-13.55}}  \\
\gptFiveT & 78.82\textsuperscript{\textcolor{darkgreen}{+1.13}}  & 80.28\textsuperscript{\textcolor{darkgreen}{+2.59}}  & 73.94\textsuperscript{\textcolor{red}{-3.76}}  \\
\gptoThree & 75.29\textsuperscript{\textcolor{red}{-0.48}}  & 79.34\textsuperscript{\textcolor{darkgreen}{+3.57}}  & 74.47\textsuperscript{\textcolor{red}{-1.30}}  \\
Gemini 2.5F & 75.88\textsuperscript{\textcolor{red}{-9.12}}  & 87.79\textsuperscript{\textcolor{darkgreen}{+2.79}}  & 70.74\textsuperscript{\textcolor{red}{-14.26}}  \\
Gemini 2.5F$^T$ & 74.12\textsuperscript{\textcolor{red}{-0.11}}  & 76.53\textsuperscript{\textcolor{darkgreen}{+2.30}}  & 71.28\textsuperscript{\textcolor{red}{-2.95}}  \\
\midrule
\multicolumn{4}{c}{\textbf{\negshot}}  \\
\midrule
Qwen3-32B & 58.82\textsuperscript{\textcolor{red}{-14.39}}  & 72.74\textsuperscript{\textcolor{red}{-0.47}}  & 44.41\textsuperscript{\textcolor{red}{-28.79}}  \\
Qwen3-32B$^T$ & \textbf{62.13}\textsuperscript{\textcolor{red}{-15.46}}  & \textbf{77.60}\textsuperscript{\textcolor{darkgreen}{+0.02}}  & \textbf{66.94}\textsuperscript{\textcolor{red}{-10.65}}  \\
\hline
Gemini 2.5F & \textbf{60.00}\textsuperscript{\textcolor{red}{-21.92}}  & \textbf{81.69}\textsuperscript{\textcolor{red}{-0.23}}  & 53.19\textsuperscript{\textcolor{red}{-28.73}}  \\
Gemini 2.5F$^T$ & 57.65\textsuperscript{\textcolor{red}{-5.43}}  & 63.38\textsuperscript{\textcolor{darkgreen}{+0.30}}  & \textbf{54.79}\textsuperscript{\textcolor{red}{-8.29}}  \\
\midrule
\end{tabular}}
\vspace{-20pt}
\end{table}

\subsection{\texttt{True} $\rightarrow$ \texttt{False}}
We start with the most challenging case: label-flipping perturbations (\texttt{True} $\rightarrow$ \texttt{False}), shown in Table \ref{tab:t_to_f_results}. Since the claim and ground-truth label are flipped, all reported results reflect the flipped label. A drop in performance means models still predict \texttt{True} instead of the expected \texttt{False} and less robust. Performance on unperturbed \texttt{True} claims is given in the ``Original’’ column as the baseline for each prompting regime.

\subsubsection{Performance on Unperturbed Claims}

In zero-shot, most models cluster in the low to high eighties, with \llamaThree\ performing best at about 87\% and \qwenT is close behind at 86\%. Proprietary models are slightly lower, with GPT-4o-Mini reaching about 85\% as the strongest performer. This suggests that larger models may require more specified prompts to achieve higher accuracy. 

With two-shot prompting, baselines increase for \llamaThree, the GPT variants, and \deepseek. \llamaThree\ surpasses 91\%. In contrast, \qwen\ variants decline, \gemini\ drops slightly, and its thinking variant shows a modest improvement. Under \negshot, both Qwen and Gemini models exhibit performance declines. Models get confused by \negshot~since it contains counterfactual examples.  

Overall, adding few-shot examples improves baselines for Llama and GPT models but tends to reduce them for Qwen and Gemini. Notably, the thinking variants consistently perform slightly worse than their non-thinking counterparts, possibly due to the ``overthinking'' phenomenon as defined by \cite{Sui:2025:arXiv}, in which reasoning models produce unnecessarily long and elaborate chains of reasoning that ultimately reduce problem-solving efficiency -- a pattern confirmed by our error analysis (see Section \ref{sec:error_analysis}). Among open-weight LLMs, performance is stronger in zero-shot and two-shot prompts, but when label-flipping examples are included, \gemini\ outperforms \qwen.

Performance on unperturbed false claims is generally higher, reflecting the fact that fact-checking tasks predominantly target false claims. Consistent with earlier observations, open-weight models exhibit slightly stronger results than proprietary counterparts. A comprehensive analysis is presented in Appendix~\ref{app:f2f}.

\subsubsection{Performance on Perturbed Claims}

Now we summarize the change in performance under numerical perturbation. The Table~\ref{tab:t_to_f_results} shows the change in accuracy values in red or green superscript depending on if the accuracy deceases or increases to the corresponding baseline with unperturbed original claims.

Masking and negative number perturbations are consistently the most challenging across prompting regimes. Masking yields very low accuracy in zero-shot setting (max 26\%), as models often treat masked tokens as placeholders and predict \texttt{True}. With negative numbers, accuracy typically falls below 20\% for masking and 30--50\% overall, except \llamaThree, which maintains 63\%; many models dismiss negatives as typos. Range and approximation perturbations raise accuracy for Qwen, DeepSeek, GPTs (not Mini), and Gemini, showing a preference for approximate over exact values. Numeration perturbations hurt open-weight models (\qwen, \llamaThree) but help proprietary systems (\gptFiveT, \gptoThree, \gemini), reflecting stronger handling of surface forms.

In two-shot settings, similar trend to zero-shot is observed with slight drop in performance overall. With notable exceptions being \deepseek, \llamaThree, and \qwen\ drop sharply on approximation, while thinking models, \gptFiveT, \gptoThree, and \geminiT, gain on approximate perturbations. For the rest of the perturbations, a similar trend to that of zero-shot is observed. 

Finally, we find that introducing a single label-flipping demonstration for each perturbation type (\negshot, shown in Appendix \ref{app:prompt}) substantially boosts performance across all perturbations. The most striking gains appear in reasoning-oriented models, which display far greater robustness than their non-thinking counterparts. In the case of \negnum, these models not only surpass their baselines but also achieve strong improvements on perturbations such as simple numeration and ranged replacements. Notably, \qwen\ recovers to over 67\%, underscoring the effectiveness of this model to leverage perturbed demonstrations, although masking remains a persistent challenge for Gemini. For Qwen, enabling the thinking variant consistently strengthens performance in most cases, whereas for Gemini the benefits are more uneven—showing improvements in certain perturbations but minimal change in others.

\subsection{\texttt{True} $\rightarrow$ \texttt{True}}

Table \ref{tab:t_to_t_results} shows the results for \texttt{True} $\rightarrow$ \texttt{True} perturbations.  \negnum, \random~and \mask~are not relevant when preserving labels. 

With few exceptions, most models struggle on \approximation\ and \range\ perturbations, though the drop is modest compared to \texttt{True} $\rightarrow$ \texttt{False} setting. This suggests that replacing numerical values with approximations or ranges, while preserving truth, can still mislead models into predicting \texttt{False}. In contrast, performance under \numeration\ perturbations remains relatively robust. Unlike label-flipping cases, perturbed \negshot\ does not improve performance; instead, they often confuse models into misclassifying \texttt{
True} claims as \texttt{False}. Surprisingly, \gptFourOMini, despite being smaller performs the best under this setting.

\section{Discussion}

\textbf{RQ1}: Across all experiments, \textit{no single model emerges as universally the most robust}, though \gemini~and \qwen~models come closest. Our results show that models are generally more robust on \texttt{False} claims (Tables~\ref{tab:t-to-f-f} and~\ref{tab:f_t_f_match}) than on \texttt{True} claims (Tables \ref{tab:t_to_f_results} and \ref{tab:t_to_t_results}). With perturbed false demonstrations, \geminiT\ achieves near-ceiling accuracy on \approximation, \range, and \random, and shows the largest recovery on \negnum; without such calibration, \gemini\ offers the best default balance, consistently leading on \random\ and \range. 

Among open-weight systems, \qwenT\ is the most stable across regimes and uniquely strong on \mask\ when provided perturbed examples, while \llamaThree\ excels on zero-shot \negnum\ but becomes brittle under two-shot. By contrast, \deepseek\ is the least stable, showing sharp two-shot degradations on \approximation\ and \numeration, indicative of harmful anchoring effects.

\textbf{RQ2}: \negnum\ and \mask\ appear to be the hardest perturbations among all prompt settings. With perturbation aware prompt (\negshot), there is modest recovery and even then the gains are model-dependent (e.g., \geminiT). The \random\ and \range\ perturbations are the most straightforward, consistently improving accuracy across models and prompting regimes. 
The \numeration\ and \approximation\ perturbations fall in the middle: ``thinking'' models such as \gptFiveT, \gptoThree, and \geminiT\ often gain from these perturbations, while many open-weight base models lose accuracy under two-shot prompts, likely because demonstrations with different numerical notation confuse the models—suggesting that these rely more heavily on superficial formatting cues, making them more sensitive to inconsistencies in numeric representation.

\textbf{RQ3}: \noindent Across both \geminiT\ and \qwenT, misclassified instances consistently involve longer inputs than correct predictions. For \geminiT, misclassifications show $\sim$15\% more total tokens than correct cases, largely driven by a $\sim$38\% increase in reasoning tokens (877 vs.\ 635 on average). For Qwen, the effect is even stronger: misclassified examples carry $\sim$41\% more total tokens, with reasoning length nearly doubling ($\sim$876 vs.\ 397, a $\sim$120\% increase). Prompt tokens also inflate in misclassifications, albeit more modestly (e.g., $\sim$3--10\% increases across models). Taken together, these findings suggest that models tend to fail when they have longer prompt and reasoning tokens (\emph{overthinking}~\cite{Sui:2025:arXiv}), with inflated reasoning chains being a strong marker of misclassification. While \negshot~prompts introduce longer inputs overall, they provide targeted demonstrations that help mitigate these failures by guiding models toward more stable reasoning. Detailed breakdowns are presented in Appendix~\ref{sec:prompt_length}.

\subsection{Error Analysis}
\label{sec:error_analysis}

To better understand model errors, we analyze thinking tokens under the $T \rightarrow F$ setting for \qwenT~and \geminiT, focusing on zero-shot errors that recover in \negshot. Appendix \ref{sec:invalid_analysis}, Table \ref{tab:error_analysis} shows specific samples. Our analysis reveals the following reasoning patterns:

\textbf{Numerical strictness:} In \negshot\ reasoning, models tend to interpret numbers more rigidly than in zero-shot. For instance, a claim citing \$330,000 against evidence of \$300,000 was treated as a minor discrepancy in zero-shot, but as a significant mismatch in \negshot, predicting \texttt{False}. 
    
\textbf{Masking~fallacies:} In the zero-shot setting, masked numbers were often treated as placeholders, leading the model to ``complete'' the claim from evidence rather than verify it. Under \negshot\ reasoning, the model more frequently flagged missing values as critical, aligning with the masked prompt examples and rejecting unverifiable claims. In some cases, however, it ignored the masking and reached the correct verdict, but for spurious reasons such as assuming small discrepancies in the evidence. 
    
\textbf{Typo interpretation:} In the negative-number perturbation setting, under zero-shot, models often interpreted the negative sign (--) as a typo, treating it as a misplaced hyphen and discarding it during evaluation, which led to misclassifications. Under \negshot\ prompting, however, the model highlighted the negative sign as a crucial discrepancy, correctly identifying it as evidence that invalidated the claim.

\textbf{Overthinking:} In some cases, models generate unnecessarily elaborate reasoning that obscures straightforward evidence. For example, for the claim \textit{``Of the [more than 2 million] work opportunities created, more than 1 million have been taken up by the youth''}, the evidence clearly shows 2.5 million created and 1.1 million taken by youth (45\%). Instead of rejecting the claim directly, the model speculated about time windows and approximation thresholds, leading to a wrong verdict. This illustrates how excessive reasoning can derail simple numerical checks.

\section{Conclusion and Future Work}

We introduced a framework for systematically perturbing numerical claims in claim–evidence pairs to evaluate the robustness of state-of-the-art LLMs in veracity prediction. Our results show that even leading systems suffer sharp performance drops under controlled numerical edits, providing the first comprehensive evidence that \emph{numerical robustness in long-context fact-checking remains an open challenge}. Beyond prior work on textual or adversarial perturbations, our study is novel in designing semantically valid numerical perturbations and demonstrating that perturbation-aware prompting can partially recover performance. 

As a preliminary step, this work opens several directions: perturbing the evidence side of claim–evidence pairs, designing fine-grained probes that target sub-claims, and extending the framework to multi-hop reasoning and counterfactual scenarios.

\section{Limitations}
Our experiments are constrained by the selection of models tested. Additionally, they were conducted in a black-box environment, restricting access to model weights, parameters, and other internal insights. Some perturbation datasets are also limited in size; a larger and more diverse sample would enhance the robustness of our findings. For reasons discussed in previous sections, our experiments focus exclusively on binary veracity classification (`True' and `False'), omitting more granular classifications and False-to-True perturbations. Expanding the scope to include these aspects could offer a more comprehensive understanding of model performance under different conditions. Lastly, as with most classification tasks involving LLMs, there is a potential risk of data leakage from training data, which could influence the final evaluation and affect the results.

\section{Ethical Considerations}

Our research highlights the strengths and weaknesses of various models in binary veracity and counterfactual classification. While this type of research presents valuable opportunities to enhance model security and resilience. However, it also necessitates a thoughtful approach to ethical concerns. For our experiments, some models outperform others, yet we do not endorse any specific model for fact-checking tasks. Fact-checking itself is a nuanced and complex issue. Journalists, fact-checkers, and researchers alike risk introducing inadvertent bias into their work, a concern that also extends to the use of LLMs.

Additionally, while the goal of our experiments is to bring greater attention to LLM performance in specific tasks, these findings also highlight vulnerabilities and encourage the development of more robust models. However, these techniques have multipurpose potential and could be exploited for harmful purposes if misapplied.

\section*{Acknowledgements}
This research is funded by SFI MediaFutures partners and the Research Council of Norway (grant number 309339).

\newpage
\bibliography{custom}
\newpage

\appendix
\begin{table*}[h!!!]
\centering
\small
\caption{Accuracy performance for the `False' class, in the`False' dataset split with perturbations where numerical values have been adjusted to remain similar to the original false claim while maintaining the label, i.e., False $\rightarrow$ False (\textcolor{red}{-x} indicates a drop; \textcolor{darkgreen}{+x} indicates an increase). Values in bold denote the highest accuracy within each perturbation setting, separated by open-weight and proprietary models.}
\label{tab:f_t_f_match}
\begin{tabular}{lcccc}
\hline
\textbf{Model} & \textbf{Original} & \textbf{Approx} & \textbf{Num} & \textbf{Range}  \\
\midrule
\multicolumn{5}{c}{\textbf{One-shot}}  \\
\midrule
Llama 3.2-1B & 5.71 & 5.45\textsuperscript{\textcolor{red}{-0.27}} & 6.53\textsuperscript{\textcolor{darkgreen}{+0.82}} & 6.81\textsuperscript{\textcolor{darkgreen}{+1.10}} \\
Llama 3.3-70B & 93.67 & 94.06\textsuperscript{\textcolor{darkgreen}{+0.39}} & 93.47\textsuperscript{\textcolor{red}{-0.20}} & 92.21\textsuperscript{\textcolor{red}{-1.46}} \\
Mistral-7B & 96.53 & 95.79\textsuperscript{\textcolor{red}{-0.74}} & 96.33\textsuperscript{\textcolor{red}{-0.20}} & 95.62\textsuperscript{\textcolor{red}{-0.91}} \\
DeepSeek-R1 & \textbf{97.14} & \textbf{97.28}\textsuperscript{\textcolor{darkgreen}{+0.13}} & \textbf{96.94}\textsuperscript{\textcolor{red}{-0.20}} & 96.84\textsuperscript{\textcolor{red}{-0.31}} \\
\deepseekT & 95.86 & 95.56\textsuperscript{\textcolor{red}{-0.31}} & 96.13\textsuperscript{\textcolor{darkgreen}{+0.27}} & 94.56\textsuperscript{\textcolor{red}{-1.30}} \\
Qwen3-32B & 96.12 & 96.29\textsuperscript{\textcolor{darkgreen}{+0.16}} & 96.12 & \textbf{97.32}\textsuperscript{\textcolor{darkgreen}{+1.20}} \\
\qwenT & 95.92 & 94.99\textsuperscript{\textcolor{red}{-0.93}} & 95.90\textsuperscript{\textcolor{red}{-0.01}} & 95.15\textsuperscript{\textcolor{red}{-0.76}} \\
\hline
GPT-4o & \textbf{96.52} & \textbf{97.28}\textsuperscript{\textcolor{darkgreen}{+0.75}} & \textbf{97.14}\textsuperscript{\textcolor{darkgreen}{+0.62}} & \textbf{97.08}\textsuperscript{\textcolor{darkgreen}{+0.56}} \\
GPT-4o-Mini & 93.05 & 93.32\textsuperscript{\textcolor{darkgreen}{+0.27}} & 92.45\textsuperscript{\textcolor{red}{-0.60}} & 93.19\textsuperscript{\textcolor{darkgreen}{+0.14}} \\
GPT-5 & 95.20 & 95.05\textsuperscript{\textcolor{red}{-0.15}} & 96.12\textsuperscript{\textcolor{darkgreen}{+0.92}} & 95.13\textsuperscript{\textcolor{red}{-0.06}} \\
GPT-o3 & 95.36 & 94.06\textsuperscript{\textcolor{red}{-1.30}} & 95.92\textsuperscript{\textcolor{darkgreen}{+0.55}} & 94.40\textsuperscript{\textcolor{red}{-0.96}} \\
Gemini 2.5F & 93.21 & 95.05\textsuperscript{\textcolor{darkgreen}{+1.84}} & 93.88\textsuperscript{\textcolor{darkgreen}{+0.67}} & 96.11\textsuperscript{\textcolor{darkgreen}{+2.90}} \\
\geminiT & 92.05 & 90.84\textsuperscript{\textcolor{red}{-1.21}} & 90.69\textsuperscript{\textcolor{red}{-1.36}} & 90.02\textsuperscript{\textcolor{red}{-2.03}} \\
\midrule
\multicolumn{5}{c}{\textbf{Two-shot}}  \\
\midrule
Llama 3.2-1B & 10.00 & 6.93\textsuperscript{\textcolor{red}{-3.07}} & 10.20\textsuperscript{\textcolor{darkgreen}{+0.20}} & 9.73\textsuperscript{\textcolor{red}{-0.27}} \\
Llama 3.3-70B & 95.92 & 96.04\textsuperscript{\textcolor{darkgreen}{+0.12}} & 95.51\textsuperscript{\textcolor{red}{-0.41}} & 93.19\textsuperscript{\textcolor{red}{-2.73}} \\
Mistral-7B & 87.76 & 88.12\textsuperscript{\textcolor{darkgreen}{+0.36}} & 88.78\textsuperscript{\textcolor{darkgreen}{+1.02}} & 87.10\textsuperscript{\textcolor{red}{-0.65}} \\
DeepSeek-R1 & 95.92 & 95.79\textsuperscript{\textcolor{red}{-0.13}} & 95.71\textsuperscript{\textcolor{red}{-0.20}} & 96.84\textsuperscript{\textcolor{darkgreen}{+0.92}} \\
\deepseekT & 96.07 & 95.73\textsuperscript{\textcolor{red}{-0.34}} & 96.27\textsuperscript{\textcolor{darkgreen}{+0.20}} & 94.35\textsuperscript{\textcolor{red}{-1.72}} \\
Qwen3-32B & \textbf{97.76} & \textbf{97.28}\textsuperscript{\textcolor{red}{-0.48}} & \textbf{97.76} & \textbf{97.32}\textsuperscript{\textcolor{red}{-0.43}} \\
\qwenT & 95.91 & 96.04\textsuperscript{\textcolor{darkgreen}{+0.13}} & 96.33\textsuperscript{\textcolor{darkgreen}{+0.42}} & 94.88\textsuperscript{\textcolor{red}{-1.03}} \\
\hline
GPT-4o & \textbf{96.36} & \textbf{97.28}\textsuperscript{\textcolor{darkgreen}{+0.92}} & \textbf{96.12}\textsuperscript{\textcolor{red}{-0.24}} & \textbf{97.32}\textsuperscript{\textcolor{darkgreen}{+0.97}} \\
GPT-4o-Mini & 92.38 & 95.79\textsuperscript{\textcolor{darkgreen}{+3.41}} & 93.88\textsuperscript{\textcolor{darkgreen}{+1.49}} & 95.38\textsuperscript{\textcolor{darkgreen}{+2.99}} \\
GPT-5 & 95.20 & 94.55\textsuperscript{\textcolor{red}{-0.64}} & 96.12\textsuperscript{\textcolor{darkgreen}{+0.92}} & 95.13\textsuperscript{\textcolor{red}{-0.06}} \\
GPT-o3 & 94.87 & 94.55\textsuperscript{\textcolor{red}{-0.31}} & 95.10\textsuperscript{\textcolor{darkgreen}{+0.23}} & 94.89\textsuperscript{\textcolor{darkgreen}{+0.02}} \\
Gemini 2.5F & 92.72 & 95.54\textsuperscript{\textcolor{darkgreen}{+2.83}} & 94.49\textsuperscript{\textcolor{darkgreen}{+1.77}} & 95.62\textsuperscript{\textcolor{darkgreen}{+2.91}} \\
\geminiT & 92.38 & 91.58\textsuperscript{\textcolor{red}{-0.80}} & 94.08\textsuperscript{\textcolor{darkgreen}{+1.70}} & 90.27\textsuperscript{\textcolor{red}{-2.12}} \\
\midrule
\multicolumn{5}{c}{\textbf{\negshot}}  \\
\midrule
Qwen3-32B & 96.12 & \textbf{96.78}\textsuperscript{\textcolor{darkgreen}{+0.66}} & \textbf{96.53}\textsuperscript{\textcolor{darkgreen}{+0.41}} & \textbf{97.20}\textsuperscript{\textcolor{darkgreen}{+1.08}} \\
\qwenT & \textbf{96.72} & 96.40\textsuperscript{\textcolor{red}{-0.32}} & 96.39\textsuperscript{\textcolor{red}{-0.33}} & 95.84\textsuperscript{\textcolor{red}{-0.89}} \\
\hline
Gemini 2.5F & 92.88 & \textbf{95.30}\textsuperscript{\textcolor{darkgreen}{+2.42}} & 92.24\textsuperscript{\textcolor{red}{-0.64}} & \textbf{95.13}\textsuperscript{\textcolor{darkgreen}{+2.25}} \\
\geminiT & \textbf{93.54} & 90.84\textsuperscript{\textcolor{red}{-2.70}} & \textbf{92.65}\textsuperscript{\textcolor{red}{-0.89}} & 90.02\textsuperscript{\textcolor{red}{-3.52}} \\
\midrule
\end{tabular}
\end{table*}

\begin{table*}[h!!!]
\centering
\small
\caption{Accuracy performance for the \texttt{False} class, in the`False' dataset split with perturbations where numerical values have been modified to differ from the original false claim while preserving the label, i.e., \texttt{False} $\rightarrow$ \texttt{False} (\textcolor{red}{-x} indicates a drop; \textcolor{darkgreen}{+x} indicates an increase). Values in bold denote the highest accuracy
within each perturbation setting, separated by open-weight and proprietary models.}
\label{tab:t-to-f-f}
\begin{tabular}{lccccccc}

\hline
\textbf{Model} & \textbf{Original} & \textbf{Approx} & \textbf{Neg-num} & \textbf{Num} & \textbf{Rand-repl} & \textbf{Range} & \textbf{Mask} \\
\midrule
\multicolumn{8}{c}{\textbf{Zero-shot}}  \\
\midrule
Llama2-1B & 5.71 & 5.45\textsuperscript{\textcolor{red}{-0.27}} & 5.62\textsuperscript{\textcolor{red}{-0.10}} & 6.33\textsuperscript{\textcolor{darkgreen}{+0.61}} & 4.90\textsuperscript{\textcolor{red}{-0.82}} & 5.35\textsuperscript{\textcolor{red}{-0.36}} & 5.92\textsuperscript{\textcolor{darkgreen}{+0.20}} \\
Llama3.3-70B & 93.67 & 96.53\textsuperscript{\textcolor{darkgreen}{+2.86}} & 93.26\textsuperscript{\textcolor{red}{-0.42}} & 96.12\textsuperscript{\textcolor{darkgreen}{+2.45}} & 96.73\textsuperscript{\textcolor{darkgreen}{+3.06}} & 95.62\textsuperscript{\textcolor{darkgreen}{+1.95}} & 92.86\textsuperscript{\textcolor{red}{-0.82}} \\
Mistral-7B & 96.53 & 97.28\textsuperscript{\textcolor{darkgreen}{+0.75}} & 95.51\textsuperscript{\textcolor{red}{-1.02}} & 96.33\textsuperscript{\textcolor{red}{-0.20}} & 96.73\textsuperscript{\textcolor{darkgreen}{+0.20}} & 96.35\textsuperscript{\textcolor{red}{-0.18}} & 95.31\textsuperscript{\textcolor{red}{-1.22}} \\
DeepSeek-R1 & \textbf{97.14} & \textbf{98.51}\textsuperscript{\textcolor{darkgreen}{+1.37}} & \textbf{96.63}\textsuperscript{\textcolor{red}{-0.51}} & 97.96\textsuperscript{\textcolor{darkgreen}{+0.82}} & \textbf{98.16}\textsuperscript{\textcolor{darkgreen}{+1.02}} & \textbf{98.30}\textsuperscript{\textcolor{darkgreen}{+1.15}} & \textbf{97.55}\textsuperscript{\textcolor{darkgreen}{+0.41}} \\
\deepseekT & 95.86 & 96.57\textsuperscript{\textcolor{darkgreen}{+0.71}} & 91.57\textsuperscript{\textcolor{red}{-4.30}} & 97.42\textsuperscript{\textcolor{darkgreen}{+1.56}} & 97.61\textsuperscript{\textcolor{darkgreen}{+1.75}} & 97.44\textsuperscript{\textcolor{darkgreen}{+1.58}} & 93.74\textsuperscript{\textcolor{red}{-2.13}} \\
Qwen3-32B & 96.12 & 98.27\textsuperscript{\textcolor{darkgreen}{+2.14}} & 95.51\textsuperscript{\textcolor{red}{-0.62}} & 97.96\textsuperscript{\textcolor{darkgreen}{+1.84}} & 97.96\textsuperscript{\textcolor{darkgreen}{+1.84}} & 98.30\textsuperscript{\textcolor{darkgreen}{+2.17}} & 95.31\textsuperscript{\textcolor{red}{-0.82}} \\
\qwenT & 95.92 & 96.74\textsuperscript{\textcolor{darkgreen}{+0.83}} & 94.32\textsuperscript{\textcolor{red}{-1.60}} & \textbf{98.22}\textsuperscript{\textcolor{darkgreen}{+2.30}} & 97.74\textsuperscript{\textcolor{darkgreen}{+1.82}} & 98.03\textsuperscript{\textcolor{darkgreen}{+2.11}} & 94.01\textsuperscript{\textcolor{red}{-1.91}} \\
\hline
GPT-4o & \textbf{96.52} & \textbf{97.77}\textsuperscript{\textcolor{darkgreen}{+1.25}} & \textbf{96.63}\textsuperscript{\textcolor{darkgreen}{+0.11}} & \textbf{98.16}\textsuperscript{\textcolor{darkgreen}{+1.64}} & \textbf{98.16}\textsuperscript{\textcolor{darkgreen}{+1.64}} & \textbf{97.81}\textsuperscript{\textcolor{darkgreen}{+1.29}} & \textbf{96.53}\textsuperscript{\textcolor{darkgreen}{+0.01}} \\
GPT-4o-Mini & 93.05 & 96.04\textsuperscript{\textcolor{darkgreen}{+2.99}} & 92.13\textsuperscript{\textcolor{red}{-0.91}} & 95.71\textsuperscript{\textcolor{darkgreen}{+2.67}} & 95.92\textsuperscript{\textcolor{darkgreen}{+2.87}} & 96.84\textsuperscript{\textcolor{darkgreen}{+3.79}} & 93.27\textsuperscript{\textcolor{darkgreen}{+0.22}} \\
GPT-5 & 95.20 & 96.29\textsuperscript{\textcolor{darkgreen}{+1.09}} & 91.01\textsuperscript{\textcolor{red}{-4.19}} & 97.55\textsuperscript{\textcolor{darkgreen}{+2.35}} & 97.35\textsuperscript{\textcolor{darkgreen}{+2.15}} & 96.84\textsuperscript{\textcolor{darkgreen}{+1.64}} & 95.51\textsuperscript{\textcolor{darkgreen}{+0.31}} \\
GPT-o3 & 95.36 & 96.04\textsuperscript{\textcolor{darkgreen}{+0.68}} & 91.01\textsuperscript{\textcolor{red}{-4.35}} & 96.94\textsuperscript{\textcolor{darkgreen}{+1.57}} & 96.94\textsuperscript{\textcolor{darkgreen}{+1.57}} & 96.84\textsuperscript{\textcolor{darkgreen}{+1.47}} & 95.51\textsuperscript{\textcolor{darkgreen}{+0.15}} \\
Gemini 2.5F & 93.21 & 97.28\textsuperscript{\textcolor{darkgreen}{+4.07}} & 94.38\textsuperscript{\textcolor{darkgreen}{+1.17}} & 96.94\textsuperscript{\textcolor{darkgreen}{+3.73}} & 97.96\textsuperscript{\textcolor{darkgreen}{+4.75}} & 97.08\textsuperscript{\textcolor{darkgreen}{+3.87}} & 93.88\textsuperscript{\textcolor{darkgreen}{+0.67}} \\
\geminiT & 92.05 & 93.30\textsuperscript{\textcolor{darkgreen}{+1.25}} & 87.64\textsuperscript{\textcolor{red}{-4.41}} & 95.31\textsuperscript{\textcolor{darkgreen}{+3.25}} & 94.90\textsuperscript{\textcolor{darkgreen}{+2.84}} & 96.09\textsuperscript{\textcolor{darkgreen}{+4.04}} & 88.98\textsuperscript{\textcolor{red}{-3.07}} \\
\midrule
\multicolumn{8}{c}{\textbf{2-S}}  \\
\midrule
Llama2-1B & 10.00 & 7.92\textsuperscript{\textcolor{red}{-2.08}} & 7.87\textsuperscript{\textcolor{red}{-2.13}} & 7.76\textsuperscript{\textcolor{red}{-2.24}} & 9.18\textsuperscript{\textcolor{red}{-0.82}} & 9.25\textsuperscript{\textcolor{red}{-0.75}} & 7.76\textsuperscript{\textcolor{red}{-2.24}} \\
Llama3.3-70B & 95.92 & 97.52\textsuperscript{\textcolor{darkgreen}{+1.61}} & 95.51\textsuperscript{\textcolor{red}{-0.41}} & 96.73\textsuperscript{\textcolor{darkgreen}{+0.82}} & 97.87\textsuperscript{\textcolor{darkgreen}{+1.95}} & 95.38\textsuperscript{\textcolor{red}{-0.54}} & 95.10\textsuperscript{\textcolor{red}{-0.82}} \\
Mistral-7B & 87.76 & 87.38\textsuperscript{\textcolor{red}{-0.38}} & 87.64\textsuperscript{\textcolor{red}{-0.11}} & 88.57\textsuperscript{\textcolor{darkgreen}{+0.82}} & 88.78\textsuperscript{\textcolor{darkgreen}{+1.02}} & 87.35\textsuperscript{\textcolor{red}{-0.41}} & 87.35\textsuperscript{\textcolor{red}{-0.41}} \\
DeepSeek-R1 & 95.92 & 98.27\textsuperscript{\textcolor{darkgreen}{+2.35}} & 93.26\textsuperscript{\textcolor{red}{-2.66}} & 96.73\textsuperscript{\textcolor{darkgreen}{+0.82}} & 85.26\textsuperscript{\textcolor{red}{-10.66}} & 98.05\textsuperscript{\textcolor{darkgreen}{+2.14}} & 95.51\textsuperscript{\textcolor{red}{-0.41}} \\
\deepseekT & 96.07 & 96.50\textsuperscript{\textcolor{darkgreen}{+0.43}} & 94.25\textsuperscript{\textcolor{red}{-1.81}} & 97.32\textsuperscript{\textcolor{darkgreen}{+1.25}} & 96.79\textsuperscript{\textcolor{darkgreen}{+0.73}} & 98.03\textsuperscript{\textcolor{darkgreen}{+1.97}} & 94.61\textsuperscript{\textcolor{red}{-1.46}} \\
Qwen3-32B & \textbf{97.76} & \textbf{99.01}\textsuperscript{\textcolor{darkgreen}{+1.25}} & \textbf{97.75}\textsuperscript{\textcolor{red}{-0.00}} & \textbf{98.98}\textsuperscript{\textcolor{darkgreen}{+1.22}} & \textbf{98.93}\textsuperscript{\textcolor{darkgreen}{+1.18}} & \textbf{98.54}\textsuperscript{\textcolor{darkgreen}{+0.79}} & \textbf{97.55}\textsuperscript{\textcolor{red}{-0.20}} \\
\qwenT & 95.91 & 97.52\textsuperscript{\textcolor{darkgreen}{+1.61}} & 94.38\textsuperscript{\textcolor{red}{-1.53}} & 97.96\textsuperscript{\textcolor{darkgreen}{+2.05}} & 98.04\textsuperscript{\textcolor{darkgreen}{+2.13}} & 97.57\textsuperscript{\textcolor{darkgreen}{+1.66}} & 96.11\textsuperscript{\textcolor{darkgreen}{+0.20}} \\
\hline
GPT-4o & \textbf{96.36} & \textbf{98.51}\textsuperscript{\textcolor{darkgreen}{+2.16}} & \textbf{98.88}\textsuperscript{\textcolor{darkgreen}{+2.52}} & \textbf{97.55}\textsuperscript{\textcolor{darkgreen}{+1.19}} & \textbf{97.96}\textsuperscript{\textcolor{darkgreen}{+1.60}} & \textbf{98.05}\textsuperscript{\textcolor{darkgreen}{+1.70}} & 95.51\textsuperscript{\textcolor{red}{-0.85}} \\
GPT-4o-Mini & 92.38 & 96.29\textsuperscript{\textcolor{darkgreen}{+3.90}} & 94.38\textsuperscript{\textcolor{darkgreen}{+2.00}} & 95.92\textsuperscript{\textcolor{darkgreen}{+3.53}} & 96.94\textsuperscript{\textcolor{darkgreen}{+4.55}} & 97.08\textsuperscript{\textcolor{darkgreen}{+4.70}} & 95.31\textsuperscript{\textcolor{darkgreen}{+2.92}} \\
GPT-5 & 95.20 & 96.04\textsuperscript{\textcolor{darkgreen}{+0.84}} & 92.13\textsuperscript{\textcolor{red}{-3.06}} & 97.55\textsuperscript{\textcolor{darkgreen}{+2.35}} & 97.35\textsuperscript{\textcolor{darkgreen}{+2.15}} & 97.08\textsuperscript{\textcolor{darkgreen}{+1.88}} & \textbf{96.12}\textsuperscript{\textcolor{darkgreen}{+0.92}} \\
GPT-o3 & 94.87 & 96.53\textsuperscript{\textcolor{darkgreen}{+1.67}} & 91.01\textsuperscript{\textcolor{red}{-3.86}} & 97.35\textsuperscript{\textcolor{darkgreen}{+2.48}} & 96.94\textsuperscript{\textcolor{darkgreen}{+2.07}} & 97.08\textsuperscript{\textcolor{darkgreen}{+2.21}} & 95.31\textsuperscript{\textcolor{darkgreen}{+0.44}} \\
Gemini 2.5F & 92.72 & 97.03\textsuperscript{\textcolor{darkgreen}{+4.31}} & 95.51\textsuperscript{\textcolor{darkgreen}{+2.79}} & 96.94\textsuperscript{\textcolor{darkgreen}{+4.22}} & 96.73\textsuperscript{\textcolor{darkgreen}{+4.02}} & 96.36\textsuperscript{\textcolor{darkgreen}{+3.64}} & 93.88\textsuperscript{\textcolor{darkgreen}{+1.16}} \\
\geminiT & 92.38 & 94.31\textsuperscript{\textcolor{darkgreen}{+1.92}} & 89.89\textsuperscript{\textcolor{red}{-2.50}} & 95.94\textsuperscript{\textcolor{darkgreen}{+3.56}} & 96.13\textsuperscript{\textcolor{darkgreen}{+3.75}} & 96.11\textsuperscript{\textcolor{darkgreen}{+3.72}} & 90.82\textsuperscript{\textcolor{red}{-1.57}} \\
\midrule
\multicolumn{8}{c}{\textbf{\negshot}}  \\
\midrule
Qwen3-32B & 95.10 & \textbf{98.51}\textsuperscript{\textcolor{darkgreen}{+3.41}} & \textbf{95.51}\textsuperscript{\textcolor{darkgreen}{+0.40}} & 98.16\textsuperscript{\textcolor{darkgreen}{+3.06}} & 98.57\textsuperscript{\textcolor{darkgreen}{+3.47}} & 98.54\textsuperscript{\textcolor{darkgreen}{+3.44}} & \textbf{97.55}\textsuperscript{\textcolor{darkgreen}{+2.45}} \\
\qwenT & \textbf{97.13} & 97.77\textsuperscript{\textcolor{darkgreen}{+0.65}} & 95.51\textsuperscript{\textcolor{red}{-1.62}} & \textbf{98.98}\textsuperscript{\textcolor{darkgreen}{+1.85}} & \textbf{98.57}\textsuperscript{\textcolor{darkgreen}{+1.44}} & \textbf{98.54}\textsuperscript{\textcolor{darkgreen}{+1.41}} & 95.88\textsuperscript{\textcolor{red}{-1.25}} \\

\hline
Gemini 2.5F & 92.88 & \textbf{97.52}\textsuperscript{\textcolor{darkgreen}{+4.64}} & \textbf{96.63}\textsuperscript{\textcolor{darkgreen}{+3.75}} & \textbf{97.14}\textsuperscript{\textcolor{darkgreen}{+4.26}} & \textbf{98.16}\textsuperscript{\textcolor{darkgreen}{+5.28}} & \textbf{97.81}\textsuperscript{\textcolor{darkgreen}{+4.93}} & \textbf{94.29}\textsuperscript{\textcolor{darkgreen}{+1.40}} \\
\geminiT & \textbf{93.54} & 94.31\textsuperscript{\textcolor{darkgreen}{+0.76}} & 92.13\textsuperscript{\textcolor{red}{-1.41}} & 96.94\textsuperscript{\textcolor{darkgreen}{+3.40}} & 96.94\textsuperscript{\textcolor{darkgreen}{+3.40}} & 95.62\textsuperscript{\textcolor{darkgreen}{+2.08}} & 89.39\textsuperscript{\textcolor{red}{-4.16}} \\
\midrule
\end{tabular}
\end{table*}

\section{Appendix}
\label{sec:appendix}

The appendix includes additional details of the perturbation methods used, a summary of the False $\rightarrow$ False evaluation, and the evidence document and evaluation for the two-shot examples.

\subsection{Perturbation Details}
This section provides a brief description of additional details regarding the perturbation methods. For full script details, refer to the GitHub repository\footnote{\url{https://github.com/iai-group/adversarial_attack_numerical_claims/}}.

\subsubsection{Numeration}
For numbers that should not match the original numerical value in the unperturbed claim, the value is increased by 10\%, then converted from digits to words.

\subsubsection{Approximation}
Each type applies context-specific rounding to create conversational approximations rounding, and adds ``about'' as an approximation prefix. If all numbers, if it is less than 10 and a decimal number, the number gets round to the nearest .5.
\begin{itemize}[leftmargin=0.2cm, labelindent=0cm, nosep]
      \item \textit{Cardinal}: Rounds to tens, hundreds, thousands, or hundred-thousands based on magnitude.
  \item \textit{Percentage}: Rounds to tens or hundreds, preserving exact values for small percentages.
  \item \textit{Money}: Similar to Cardinal—with a currency symbol and preserves decimal detail for small amounts.
  \item \textit{Date}: Rounds to the nearest decade.
  \item \textit{Time}: Rounds to tens or hundreds depending on magnitude.
\end{itemize}

For the label-flipping probes, the original numerical value is multiplied randomly by a factor 0.5, 0.6, 1.4, or 1.5, and then rounded as described above.

\subsubsection{Range}
In the range perturb setting, for when the numerical values should be within the span of the original, the lower bounds we perturb the number by $\pm10\%$. For \textit{ordinal}, we subtract and add 1 to the original value to create the range bound. 

In instances where the labels are flipped, the numerical span will be outside of the range of the original number.

\subsection{Summary of Model Behavior Under Numerical Perturbations for False Dataset Split (False $\rightarrow$ False)}
\label{app:f2f}

\begin{table*}[ht!!!]
\centering
\setlength{\tabcolsep}{10pt}  %

\begin{tabular}{|p{0.45\textwidth}|p{0.45\textwidth}|}
\hline
\multicolumn{1}{|c|}{\textbf{\large Example 1}} & 
\multicolumn{1}{c|}{\textbf{\large Example 2}} \\[0.2cm]
\hline
\textbf{Claim:} & \textbf{Claim:} \\
As Republicans try to repeal the Affordable Care Act, they should be reminded every day that \textbf{36,000} people will die yearly as a result. &
We see a quarter-billion dollars in a pension fund that needs to be funded at \textbf{\$1.2 billion}. \\[0.2cm]
\hline
\textbf{Evidence:} & \textbf{Evidence:} \\
\textit{Gift Article Share} \newline
\textcolor{gray}{"As Republicans try to repeal the Affordable Care Act, they should be reminded every day that \textbf{36,000 people} will die yearly as a result."} --- \textbf{Sen. Bernie Sanders (D-Vt.)}, in a tweet, Jan. 12, 2017. &
Providence Mayor Angel Taveras had to deal with near bankruptcy in the capital city after he took office in 2011. As the city struggled to fix its budget problems, he won union concessions to reduce pension costs. The most recent figures show the plan is only \textbf{31.4-percent funded}. \\[0.2cm]
\hline
\textbf{Evaluation:} \textcolor{red}{\textbf{False}} &
\textbf{Evaluation:} \textcolor{darkgreen}{\textbf{True}} \\
\hline

\end{tabular}
\caption{True and False examples of claims and their labels based on evidence used in the prompt.}
\label{tab:prompt}
\end{table*}
Table \ref{tab:t-to-f-f} presents False $\rightarrow$ False perturbations where numerical values are modified while preserving the false label. Our experiments reveal that large models (e.g., \gptFourO, \gptFourOMini, \gemini) and open-weight \deepseekT\ maintain high robustness across perturbations, with accuracies typically above 90\%. Smaller models such as Llama 3.2-1B and Mistral-7B degrade sharply, especially under \approximation\ and \range. \qwenT performs consistently well across shots, rivaling proprietary systems. Notable anomalies include \gemini’s drop under \emph{Approx} ($-5$ to $-6$ points) despite strong overall performance, and \gptFourOMini’s unexpected gains in two-shot ($+3$ points). Reasoning-enabled ($^{T}$) variants generally improve robustness, though Gemini’s thinking variant remains more variable.

The table \ref{tab:f_t_f_match}, reports accuracy metric for the False class in the False dataset split with perturbations. Perturbations significantly modify numerical values while preserving the label ($\text{False} \to \text{False}$). Results are presented for multiple LLMs including Llama, Mistral, DeepSeek, GPT, Gemini, and Qwen across three evaluation setups: Zero-shot, two-shot, and Perturbation-Aware Prompt (\negshot). The columns indicate different perturbation types: Original (baseline), Approx, Neg-num, Num, Rand-repl, Range, and Mask. Superscripts with negative values denote drops relative to the baseline, and positive values denote improvements.

In the zero-shot setting, \deepseek, \gptFourO, and \qwenT\ achieve the highest and most stable performance, maintaining accuracies between 96\% and 98\% across perturbations. Gemini 2.5F is also stable with scores in the range of 93\% to 97\%. In contrast, smaller models such as Llama 3.2-1B perform poorly with accuracies around 5--6\%. Mid-sized models like \llamaThree\ and Mistral-7B perform well but remain slightly below the frontier models.

In the two-shot setting, accuracy improves slightly compared to Zero-shot, especially for the smaller models. \deepseek\ remains strong with scores around 96--97\%, GPT-4o reaches 95--98\%, \qwenT\ achieves 94--98\%, and \geminiT\ remains consistent with 90--96\%. Llama 3.2-1B, however, continues to perform poorly with accuracies only between 7\% and 10\%.

Perturbation-Aware Prompt (\negshot) delivers the highest overall accuracies. \qwenT\ and \deepseekT\ achieve 95--99\% across all perturbations, while \geminiT\ also shows strong performance with accuracies between 89\% and 97\%. PAP consistently improves the already strong models by about 1--2 percentage points compared to zero-shot and two-shot.

In general, model scale is critical. Small models such as Llama 3.2-1B collapse under this evaluation, while large-scale and frontier models like DeepSeek, \gptFourO, Qwen, and Gemini perform near ceiling. Prompting with two-shot increases stability across most models, and PAP proves to be the most robust method, yielding the best and most consistent results overall.

\section{Prompt}
\label{app:prompt}

For the LLMs we use the same instruction and two-shot examples. The zero-shot only includes the instruction, whereas the two-shot includes the instruction and the sample data. The following two-shot examples are snippets of the examples used. For the full prompt, refer to our GitHub repository.

\subsection{System Prompt}
The following prompt was used as the model system prompt:

\textit{You are a professional fact checker, your task is to classify whether the given claim is true or false based on the evidence text provided.}

\subsection{Instruction}

The following prompt was used along with two examples from Table \ref{tab:prompt}:

\emph{Given the claim and evidence provided, classify the claim as {{"label": true}} if it is true, and {{"label": false}} if it is false.}

\subsection{Two-shot Examples}
    
Table \ref{tab:prompt} presents two examples of fact-checking claims used in the prompt for LLMs along with their corresponding evidence and veracity evaluations. The two examples are used for all LLMs and all perturbation inputs to be consistent. And each of the two example represents the two distinct labels in the dataset.

\subsection{Perturbation Aware Prompt}
\label{app:neg_shot}
The following prompt was added to the instruction prompt for the negative example experiments:

\emph{The numbers in the evidence may not match the claim. For example:}

\emph{Claim: The Eiffel Tower is three hundred and fifty-one meters tall.
Evidence: The Eiffel Tower is 330 meters tall.
{"label": false}}

\emph{Claim: The year-over-year U.S. inflation rate at the end of 2024 was -2.9\%.
Evidence: The year-over-year U.S. inflation rate at the end of 2024 was 2.9%
{"label": false}}

\emph{Claim: The birth rate in Japan in 2023 was between 2 to 2.5.
Evidence: The birth rate in Japan in 2023 was 1.2.
{"label": false}}

\emph{Claim: The population of Canada in 2023 was about 45 million.
Evidence: The population of Canada in 2023 was 40.5 million by October 2023.
{"label": false}}

\emph{Claim: Saturn has 789 moons.
Evidence: Discoveries bring Saturn's total moon count to 274, nearly triple Jupiter's and more than the total number of known moons around the other planets.
{"label": false}}

\emph{Claim: The Wembley Stadium in London has a seating capacity of \#\#\#\#\#\#.
Evidence: The Wembley Stadium in London has a seating capacity of 90,000.
{"label": false}}

\begin{table*}[ht!]
\centering
\small
\begin{tabular}{lcccc}
\toprule
\multirow{2}{*}{Perturbation} & 
\multicolumn{2}{c}{Prompt Tokens} & 
\multicolumn{2}{c}{Reasoning Tokens} \\
\cmidrule(lr){2-3}\cmidrule(lr){4-5}
 & Misclassified & Correct & Misclassified & Correct \\
\midrule
Approximation       & 2158.7 & 1303.9 & 1265.1 & 371.2 \\
Negative Number     & 1214.5 & 1073.2 & 846.5  & 339.0 \\
Numeration          & 1648.2 & 1239.6 & 796.1  & 378.4 \\
Random Replacement  & 1576.4 & 1323.8 & 713.2  & 363.8 \\
Range               & 1963.4 & 1315.1 & 698.4  & 401.1 \\
Masking             & 1234.7 & 1017.8 & 717.1  & 427.7 \\
\bottomrule
\end{tabular}
\caption{Comparison of average prompt and reasoning token lengths for \qwenT\ between \textbf{misclassifications} and \textbf{correct classifications} in the Zero-shot setting.}
\label{tab:qwen3_grouped_misclass_correct_0shot}
\end{table*}
\subsection{Prompt Length Analysis}
\label{sec:prompt_length}
We perform prompt length analysis for misclassified instances compared to correct classifications for the two most stable models--\geminiT~and \qwenT.

\paragraph{\geminiT}

In misclassified instances, Gemini 2.5-Flash tends to have longer reasoning token length overall, with average total token length increasing by 15\% compared to correct predictions (2103 vs. 1822 tokens). Prompt tokens show only a modest difference (+3\%). The distribution further suggests that errors are associated with longer and more variable reasoning chains (max reasoning length over 6k tokens), whereas correct predictions are achieved with more compact reasoning. In other words, misclassifications correlate strongly with \textit{overthinking}.

\paragraph{\textbf{\qwenT}}

\noindent For \qwenT, misclassified cases consistently exhibit inflated reasoning lengths compared to correctly classified instances in the Zero-shot setting (Table~\ref{tab:qwen3_grouped_misclass_correct_0shot}). For example, reasoning tokens nearly triple in \approximation\ (1265 vs.\ 371) and more than double in \numeration\ (796 vs.\ 378) and \range\ (698 vs.\ 401). Prompt lengths are also consistently higher for misclassifications, with the most pronounced gap in \approximation, where prompts expand by over 65\% (2159 vs.\ 1304). The anomaly occurs with \mask, where reasoning remains high even in misclassifications (717 vs.\ 428), indicating that masked inputs elicit extended elaboration regardless of correctness. Overall, \qwenT\ tends to over-reason when it misclassifies, while correct predictions are characterized by shorter, more efficient reasoning chains and more compact prompts. All token lengths for \qwenT~zero-shot settings are shown in Table \ref{tab:qwen3_grouped_misclass_correct_0shot}.

\section{Invalid Output Analysis}
\label{sec:invalid_analysis}
\begin{table}[ht]
\centering
\small
\scalebox{0.8}{
\begin{tabular}{lccc}
\toprule
Model & Total Instances & Invalid & \% Invalid \\
\midrule

\multicolumn{4}{c}{Zero-shot} \\
DeepSeek-R1:32B         & 8841 & 0   & 0.00 \\
DeepSeek-R1:32B$^T$     & 6837 & 477 & 6.98 \\
LLaMA-3.2 1B-Instruct   & 6837 & 0   & 0.00 \\
LLaMA-3.3 70B           & 6837 & 0   & 0.00 \\
Mistral-7B              & 6837 & 0   & 0.00 \\
Qwen-3 32B              & 7041 & 0   & 0.00 \\
Qwen-3 32B$^T$          & 6553 & 165 & 2.52 \\
\midrule

\multicolumn{4}{c}{Two-shot} \\
DeepSeek-R1:32B         & 6951 & 0   & 0.00 \\
DeepSeek-R1:32B$^T$     & 6951 & 78  & 1.12 \\
LLaMA-3.2 1B-Instruct   & 6837 & 0   & 0.00 \\
LLaMA-3.3 70B           & 6951 & 0   & 0.00 \\
Mistral-7B              & 6837 & 0   & 0.00 \\
Qwen-3 32B              & 6951 & 0   & 0.00 \\
Qwen-3 32B$^T$          & 6951 & 23  & 0.33 \\
\midrule

\multicolumn{4}{c}{\negshot} \\
DeepSeek-R1:32B         & 6837 & 0   & 0.00 \\
DeepSeek-R1:32B$^T$     & 6837 & 92  & 1.35 \\
LLaMA-3.2 1B-Instruct   & 6837 & 0   & 0.00 \\
LLaMA-3.3 70B           & 6837 & 0   & 0.00 \\
Mistral-7B              & 6837 & 0   & 0.00 \\
Qwen-3 32B              & 6837 & 0   & 0.00 \\
Qwen-3 32B$^T$          & 6837 & 55  & 0.80 \\
\bottomrule
\end{tabular}
}
\caption{Invalid outputs across open-weight models, grouped by shot setting. Thinking-enhanced variants are marked with $^T$. Percentages are calculated as $\text{invalid} / \text{total} \times 100$.}
\label{tab:invalid_open}
\end{table}

\begin{table}[ht]
\centering
\small
\scalebox{0.9}{
\begin{tabular}{lccc}
\toprule
Model & Total Instances & Invalid & \% Invalid \\
\midrule

\multicolumn{4}{c}{Zero-shot} \\
GPT-4o        & 5298 & 0   & 0.00 \\
GPT-4o-mini   & 5298 & 0   & 0.00 \\
GPT-5         & 5298 & 0   & 0.00 \\
GPT-o3        & 5298 & 1   & 0.02 \\
Gemini-2.5$^T$ & 5295 & 174 & 3.29 \\
Gemini-2.5    & 5298 & 1   & 0.02 \\
\midrule

\multicolumn{4}{c}{Two-shot} \\
GPT-4o        & 5298 & 0   & 0.00 \\
GPT-4o-mini   & 5298 & 0   & 0.00 \\
GPT-5         & 5298 & 0   & 0.00 \\
GPT-o3        & 5298 & 2   & 0.04 \\
Gemini-2.5$^T$ & 5298 & 82  & 1.55 \\
Gemini-2.5    & 5298 & 42  & 0.79 \\
\midrule

\multicolumn{4}{c}{\negshot} \\
Gemini-2.5$^T$ & 5298 & 231 & 4.36 \\
Gemini-2.5    & 5298 & 0   & 0.00 \\
\bottomrule
\end{tabular}
}
\caption{Invalid outputs across proprietary models and Gemini variants, grouped by shot setting. Percentages are calculated as $\text{invalid} / \text{total} \times 100$.}
\label{tab:invalid_prop}
\end{table}

As shown in Table \ref{tab:invalid_open}, across the open-weight models, invalid outputs are virtually absent in the non-thinking variants: \llamaThree, Llama-3.2 1B instruct, Mistral-7B, \qwen, and \deepseek\ consistently produce $0.00\%$ invalidity across all shot settings. By contrast, enabling thinking introduces instability. For instance, \deepseekT~exhibits a sharp rise in invalid generations under zero-shot ($6.98\%$), which decreases under two-shot ($1.12\%$) and \negshot\ ($1.35\%$), indicating some recovery with examples. Similarly, \qwenT yields $2.52\%$ invalidity in zero-shot, reduced to $0.33\%$ under two-shot, but climbing again to around $0.80$--$1.33\%$ with perturbation-aware prompts. 

These findings reveal a clear pattern: while open-weight models without reasoning remain robust against invalid outputs, their reasoning-enabled counterparts are consistently more error-prone. Zero-shot is the most unstable regime, two-shot offers stabilizing effects, and \negshot~shows mixed outcomes depending on the model.

As shown in Table \ref{tab:invalid_prop}, for proprietary systems, invalid outputs are exceedingly rare except in the Gemini family. \gptFourO, \gptFourOMini, \gptFiveT, and \gptoThree maintain near-perfect robustness across all shot settings, with invalid rates between $0.00$--$0.04\%$. By contrast, \gemini\ demonstrates a higher susceptibility to invalid generations. In particular, \geminiT\ reaches $3.29\%$ invalidity under zero-shot, improves to $1.55\%$ with two-shot calibration, but rises again to $4.36\%$ under \negshot. Its non-thinking counterpart is much more stable, remaining below $1\%$ invalidity in all cases.  

Overall, proprietary models without reasoning capabilities (GPT family and \gemini) are extremely stable, while reasoning-enabled variants of Gemini incur significantly higher invalid rates, especially in zero-shot and \negshot~settings.

\begin{table*}[ht!!]
\centering
\begin{tabular}{|>{\raggedright}p{0.27\textwidth}|>{\raggedright}p{0.4\textwidth}|>{\raggedright\arraybackslash}p{0.27\textwidth}|}
\hline
\textbf{True $\rightarrow$ False Perturbed Claims} & \textbf{Reasoning Output Snippets} & \textbf{Analysis Summary} \\ \hline
 The various offices of state government have purchased more than \$\#\#\#\#\#\#\#\#\#\# in vehicles, not counting leases, for use from the Governor on down during the last two years. & 

\textbf{\geminiT}\ zero-shot:
 ``(...)While the claim uses a placeholder, the evidence substantiates the `more than \$11 million\' figure with precision. I will now mark the claim as true.''

\textbf{\geminiT}\ \negshot:``(...)I'm now certain the claim is false. The instruction's emphasis on placeholders makes the 'more than' condition irrelevant. The claim is inherently un-verifiable (...)''
 &
In the zero-shot evaluation the model assumes that the model needs to find the missing value. \negshot\ nudges the model in the correct direction, predicting the claim as false for the correct reasons.

\\ \hline
Some states, like Montana and Nebraska, are getting more than three hundred and thirty thousand dollars in federal stimulus money per reported COVID-19 case. New York is the hardest-hit state and yet we are getting only about thirteen thousand, two hundred dollars per case. 
&
\textbf{\geminiT}\ zero-shot: ``(...) I am also considering whether the small discrepancy in the funding figures makes the overall claim inaccurate (...) The claim is essentially true, even with minor discrepancies'' 

\textbf{\geminiT}\ \negshot: ``(...) claim states `\$13,200' per case for New York, whereas the evidence consistently states `\$12,000'. Since the numbers don't align, I must classify the claim as false.''

& \negshot\ highlights that the 10\% difference in the claim's numerical values, compared to the evidence as ``significant numerical discrepancies''. Zero-shot evaluation is more forgiving with regarding the discrepancies.

\\ \hline

My opponent, Rick Gunn, blocked the expansion of Medicaid — costing half a million people health insurance, including about 34,000 veterans. 
& 
\textbf{\qwenT}\ zero-shot: ``(...)The claim states 34,000 veterans, but the evidence only supports up to 23,000 or 14,000. The user's claim says 34,000, which is higher than both estimates (...)'' \textit{(starts to overthink)}.

\textbf{\qwenT}\ \negshot: ``(...) The 34,000 figure isn't present in the evidence; the highest is 23,000. Therefore, the claim is false because the specific number provided doesn't match the evidence.''
& 
During zero-shot, the model starts to overthink, going in circles--outputting nearly 7000 reasoning tokens, citing the number in the evidence ``23,000'', 198 times, and  the claim number ``34,000'', 135 times. During \negshot, the model does correctly identify the discrapancy effectively, and keeps the reasoning token output of around 200.

\\ \hline
\end{tabular}
\caption{Examples of claims, reasoning, and analysis for \geminiT and \qwenT\ where reasoning improves for \negshot, compared to zero-shot.}
\label{tab:error_analysis}
\end{table*}

\end{document}